\documentclass[10pt,twocolumn,letterpaper]{article}

\usepackage{cvpr}
\usepackage{times}
\usepackage{epsfig}
\usepackage{graphicx}
\usepackage{amsmath}
\usepackage{amssymb}

\usepackage[pagebackref=true,breaklinks=true,letterpaper=true,colorlinks,bookmarks=false]{hyperref}

\cvprfinalcopy 

\def\cvprPaperID{****} 

\ifcvprfinal\fi
\begin{document}

\title{2nd Place Solutions for UG2+ Challenge 2022 - D$^{3}$Net for Mitigating Atmospheric Turbulence from Images}

\author{Sunder Ali Khowaja\\
Faculty of Engineering and Technology, University of Sindh, Pakistan\\ Tech University of Korea, Republic of Korea\\
{\tt\small sandar.ali@usindh.edu.pk, sunderali@tukorea.ac.kr}
\and
Jiseok Yoon\\
Tech University of Korea, Republic of Korea\\
{\tt\small jsyoon@tukorea.ac.kr}
\and
Ik Hyun Lee\\
Tech University of Korea, Republic of Korea\\
IKLab, Republic of Korea\\
{\tt\small ihlee@tukorea.ac.kr, iklab21@gmail.com}
}

\maketitle

\begin{abstract}
This technical report briefly introduces to the D$^{3}$Net proposed by our team "TUK-IKLAB" for Atmospheric Turbulence Mitigation in $UG2^{+}$ Challenge at CVPR 2022. In the light of test and validation results on textual images to improve text recognition performance and hot-air balloon images for image enhancement, we can say that the proposed method achieves state-of-the-art performance. Furthermore, we also provide a visual comparison with publicly available denoising, deblurring, and frame averaging methods with respect to the proposed work. The proposed method ranked 2nd on the final leader-board of the aforementioned challenge in the testing phase, respectively.  
\end{abstract}

\section{Introduction}

The atmospheric turbulence mitigation challenge at CVPR 2022 is a part of the Workshop on UG2+ Prize Challenge. The challenge addresses the problem of images acquired over long distances through atmospheric turbulences. It is a challenging task as there is no image domain model and no realistic way of acquiring input-output pairs without using wave-propagation simulation. Furthermore, the challenge is extended to that of the data generated as acquiring sequences representing real-world data along with its corresponding ground truth is an open research problem itself. The participants were asked to generate their dataset using a state-of-the-art simulator \cite{TurbSim} along with frames from the turbulence text dataset and hot-air balloon images. This task needs to remove the atmospheric turbulence effect from textual images captured in a hot weather from 300 meters away and generic turbulence images acquired through heated air. The hot-air balloon images have the size of 256x256 pixels, whereas the textual image dimensions were 440x440, respectively. Some samples of textual and hot-air balloon images are shown in Figure 1. For the dry run period, 20 sets of textual frames and 50 hot-air balloon images were provided, whereas, for the final test phase, the set of textual frames was extended to 100. The main task was to mitigate the turbulence effect from textual images to improve text recognition performance. 

\begin{figure}[t]
\begin{center}
   \includegraphics[width=\linewidth]{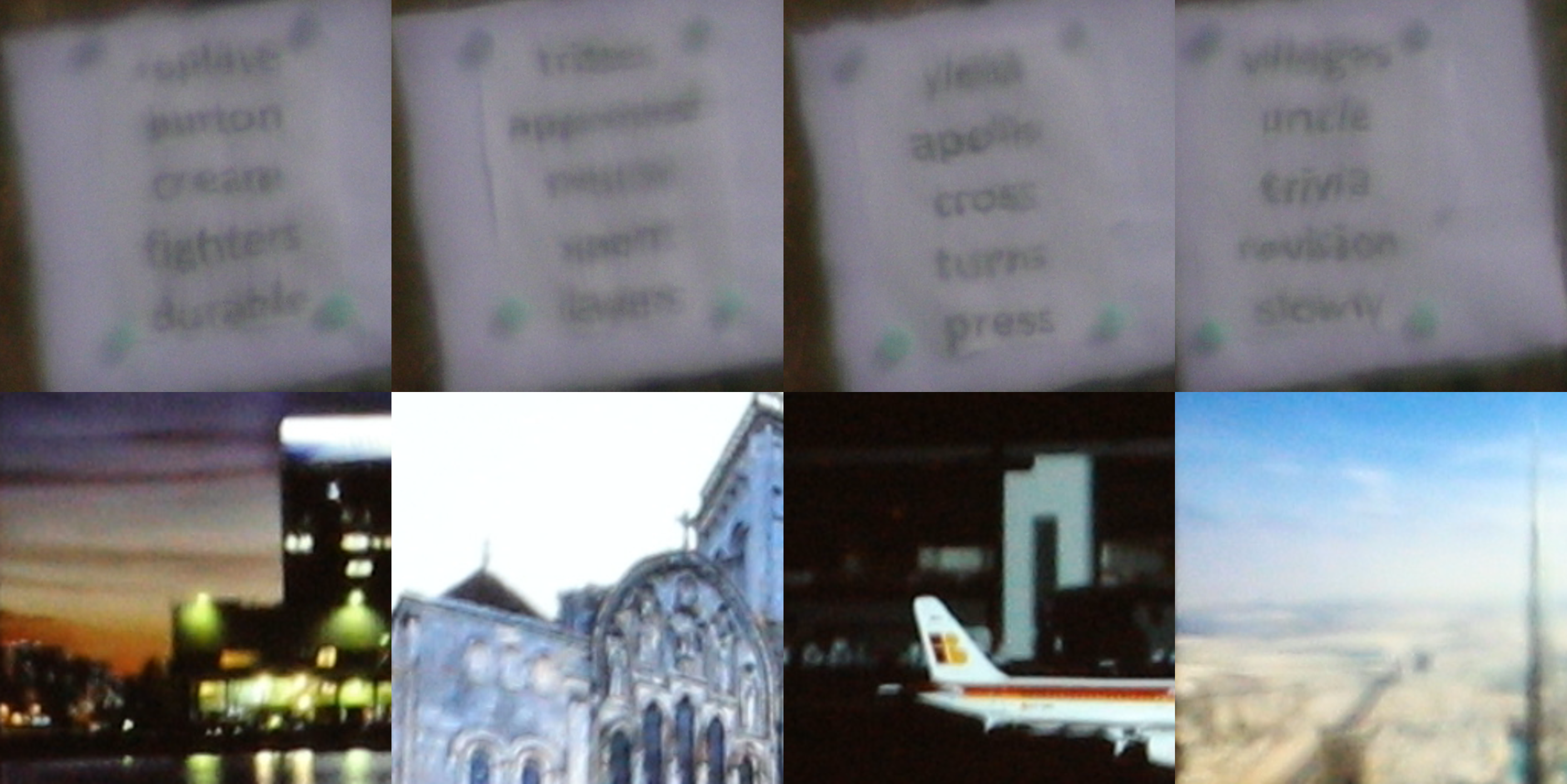}
\end{center}
   \caption{Starting frame from first four sets of textual turbulence and hot-air balloon images (testing phase)}
\label{fig:long}
\label{fig:onecol}
\end{figure}
\begin{figure*}[t]
\begin{center}
   \includegraphics[width=\linewidth]{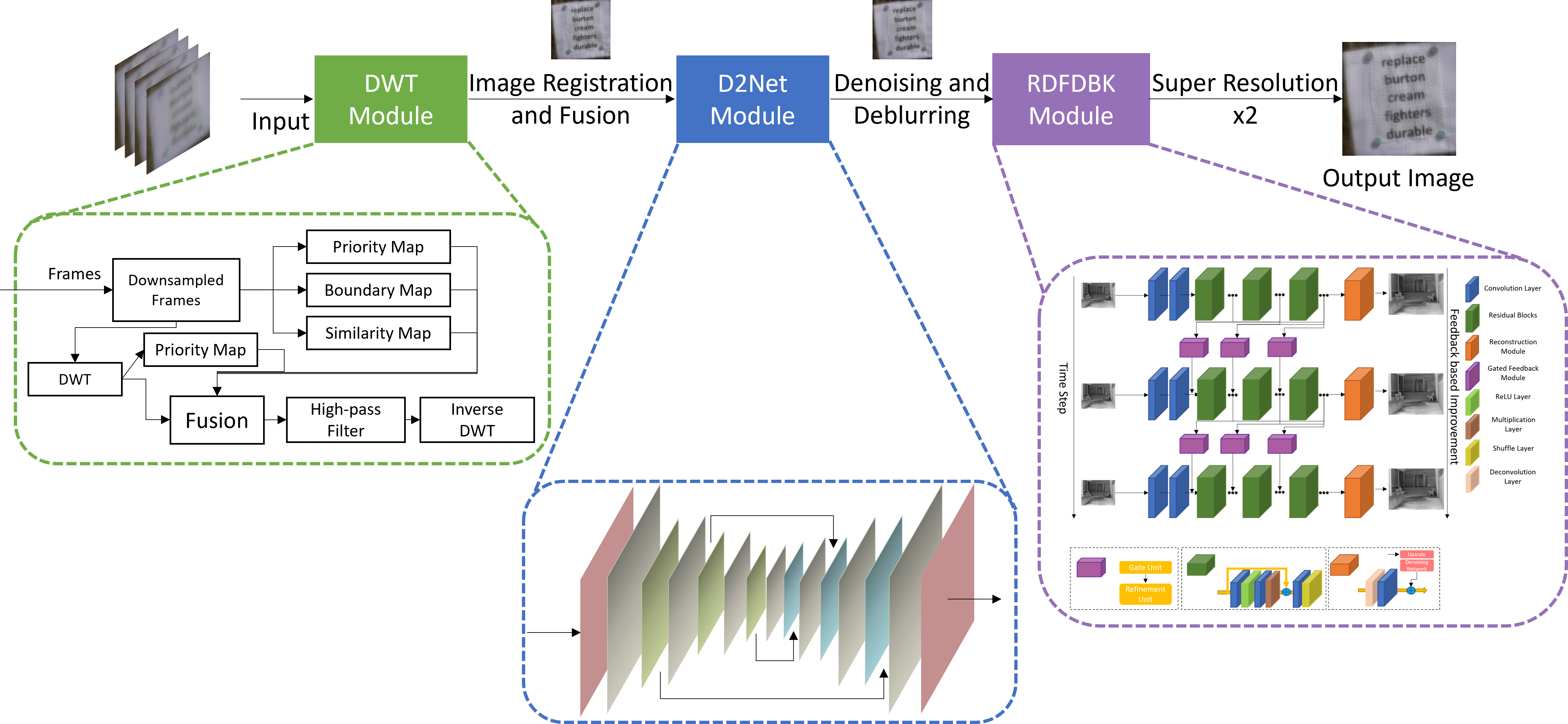}
\end{center}
   \caption{Proposed D$^{3}$ Net flow and architecture for Atmospheric Turbulence Mitigation from Images.}
\label{fig:long1}
\label{fig:onecol1}
\end{figure*}

\section{Overview of Method}

Our proposed approach is termed Discrete Wavelet, Denoising, and Deblurring Network (D$^{3}$Net) for Atmospheric Turbulence Mitigation from images. The flow and network architecture are shown in Figure 2. The network is divided into DWT, D$^{2}Net$, and RDFDBK modules. The DWT module is responsible for generating an approximate image based on priority, boundary, and similarity maps from downsampled and DWT representation-based image. The priority and boundary maps are computed from overlapping regions, whereas the similarity maps are generated using region of interest (ROI) size, intensity similarity, and sharpness. The DWT module first performs image registration and motion estimation using phase-shift properties as proposed in \cite{Chen2012}. We used six bands for extracted DWT representation-based images. The DWT module then performs a fusion of all the representation images using the weighted averaging method as proposed in \cite{Wan2009}. After the fusion, we apply inverse DWT transform to obtain the turbulence mitigated image shown in Figure 2. However, the DWT module introduces noisy artifacts and blurry effects (due to the atmospheric turbulence) that need to be removed before the recognition or enhancement phase. In this regard, we propose a Denoising and Deblurring Network (D$^{2}Net$) that leverages the characteristics of U-Net \cite{U-Net2015} and deep convolutional neural networks (CNN) to remove noise and blur artifacts. We trained D$^{2}Net$ on the DIV2K dataset \cite{DIV2k} with the proposed degradation function and then fine-tuned it on the data acquired from the TurbulenceSIM-P2S simulator \cite{TurbSim}. 

\subsection{Network Configuration}

The details for the D$^{2}Net$ are as follows: U-Net is used as a backbone network for D$^{2}Net$Net with four scales. We used 2x2 identity skip connections between strided convolution and transposed convolutional layers. The number of channels was selected to be 64, 128, 256, and 512 for each of the four scales. The residual layers were also used in the D$^{2}Net$, followed by the ReLU activation layer. The L1 loss was minimized by using an ADAM optimizer. We used the learning rate of 0.0004 and decreased the rate by 0.00005 after every 1000 iterations. We used the patch size of 16 and a batch size of 32 accordingly. The network was trained on Nvidia RTX 3060 Ti.
We use the RDFDBK module to super resolve the image for the last part. The reason for performing the image's super resolution is due to the downsampling which was performed in the DWT module. We use the residual feedback (RDFDBK) proposed by us in the NTIRE Video Super Resolution Challenge (x4SR) in CVPR 2022 \cite{NTIRE2022}. The network improves the quality of super resolution by applying feedback and sharing weights among residual layers across time steps.

\subsection{Training Data}

The training images were crawled from a license-free website\footnote{https://pixabay.com/}. Then, we selected 50 images and added the varying turbulence effect to create 5000 images (100 images for each base image) using the provided simulator TurbulenceSIM-P2S \cite{TurbSim}. In the process of generating turbulence images, we used the similar parameters suggested in the competition guidelines (“Documentation for UG2 Challenge 2022 Track 3: Atmospheric Turbulence Mitigation” – Page 4). Furthermore, we also used the images from the DIV2K dataset \cite{DIV2k} with deblurring and denoising degradation functions to train the network. 

\begin{figure}[t]
\begin{center}
   \includegraphics[width=\linewidth]{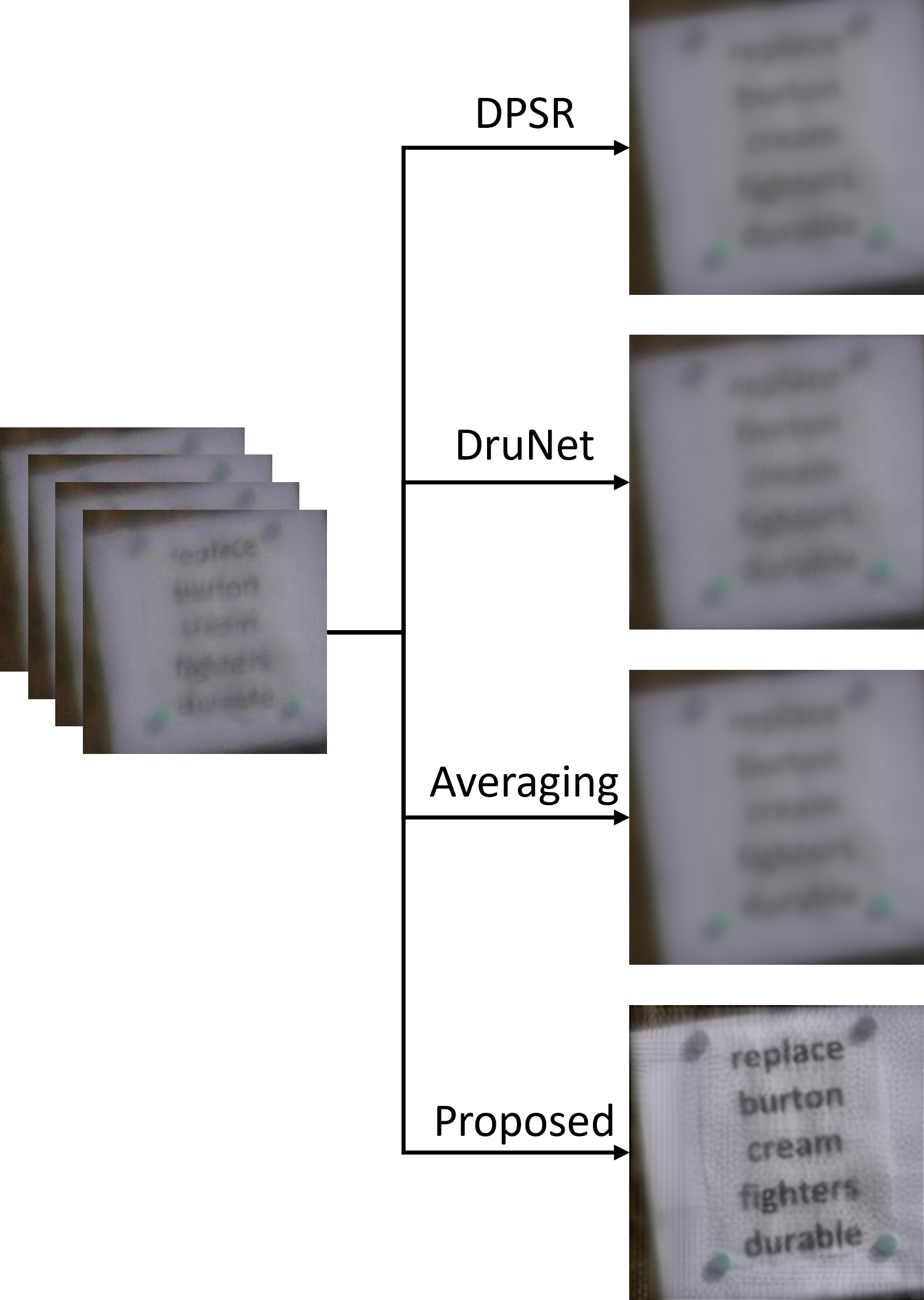}
\end{center}
   \caption{Visual comparison of the proposed method with DPSR, DruNet, and frame averaging method on a set of textual images.}
\label{fig:long2}
\label{fig:onecol2}
\end{figure}

\begin{figure}[t]
\begin{center}
   \includegraphics[width=\linewidth]{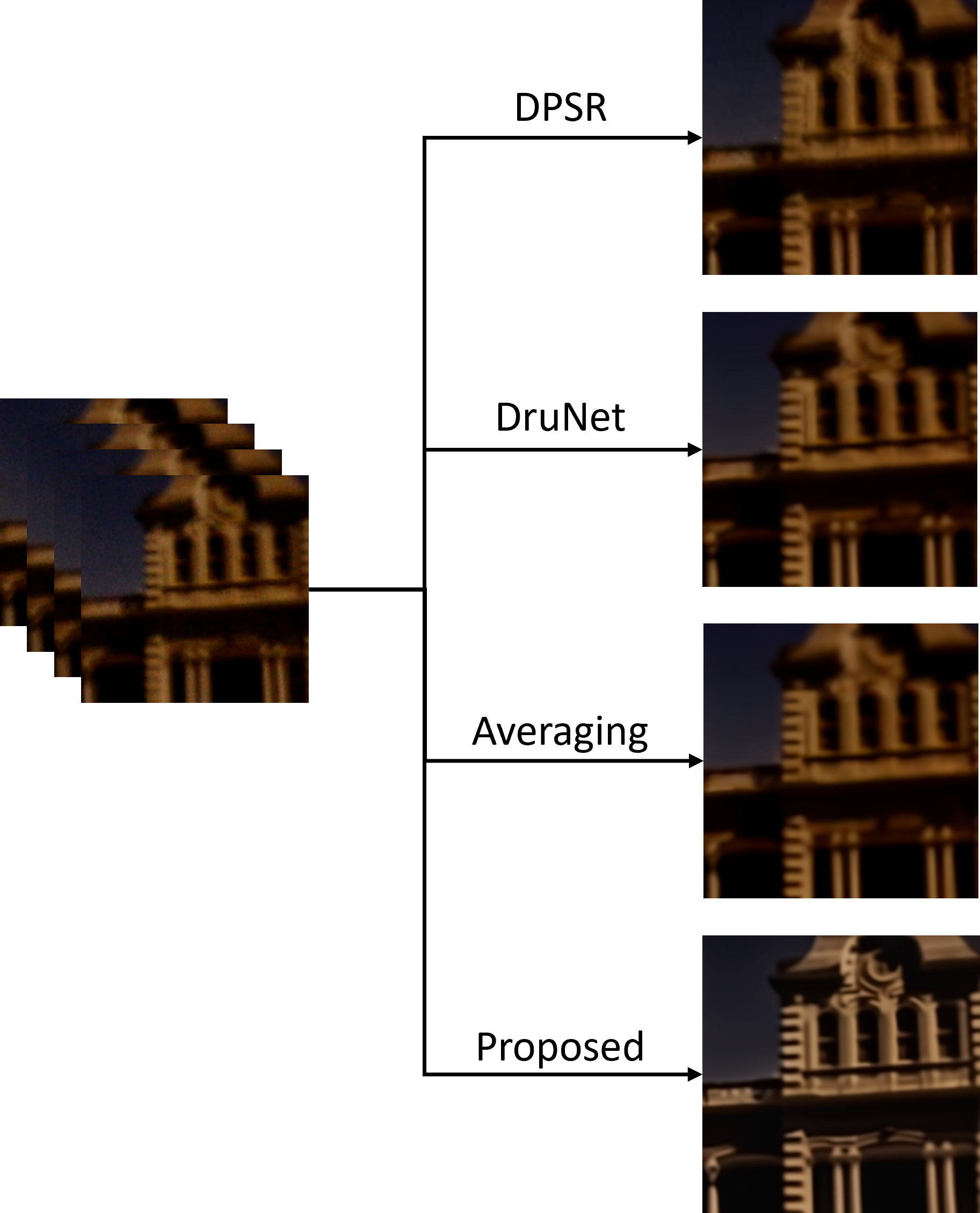}
\end{center}
   \caption{Visual comparison of the proposed method with DPSR, DruNet, and frame averaging method on a set of hot-air balloon images.}
\label{fig:long3}
\label{fig:onecol3}
\end{figure}

\section{Preliminary Results}

We show some preliminary results on both textual images and hot-air balloon images using the proposed D$^{3}Net$ in Figure 3 and 4, respectively. We also provide a comparison with DPSR \cite{DPSR}, DruNet \cite{DPIR}, and the frame averaging method from a visual perspective. We used a denoising network (DPSR) and a deblurring network (DruNet) as the authors of these networks have publicly released the pre-trained networks. There is not much difference when using the denoising, deblurring, and frame averaging method from a visual perspective. However, a drastic visual improvement could be observed compared to the proposed method.  

\section{Conclusion}
Our submission to the Atmospheric Turbulence Mitigation in $UG2^{+}$ Challenge at CVPR 2022 proposed D$^{3}$Net. The proposed network comprises of three modules, i.e., DWT, D$^{2}$Net, and RDFDBK. The network reconstructs and enhances textual and hot-air balloon images affected by atmospheric turbulence and hot weather. Our submitted results and 2nd position in the competition show that the proposed method achieves state-of-the-art results for mitigating atmospheric turbulence from images. We intend to extend our analysis to more real-world images and compare our work quantitatively with other state-of-the-art networks to validate the D$^{3}$Net's performance. 

{\small
\bibliographystyle{ieee}
\bibliography{egbib}
}

\end{document}